\pdfoutput=1

\documentclass[11pt]{article}

\usepackage{emnlp2021}

\usepackage{times}
\usepackage{latexsym}

\usepackage[T1]{fontenc}

\usepackage[utf8]{inputenc}

\usepackage{microtype}

\usepackage{url}
\usepackage{amsmath,amssymb}
\usepackage{multirow,booktabs}
\usepackage{makecell,graphicx}
\usepackage{tablefootnote}

\title{CCPM: A Chinese Classical Poetry Matching Dataset}

\author{
Wenhao Li$^{1,2}$,
Fanchao Qi$^{1,2}$,
Maosong Sun$^{1,2,3}$\thanks{\ \ Corresponding Author}\hspace{0.2em},
Xiaoyuan Yi$^{1,2}$,
Jiarui Zhang$^{2,4}$
\\ 
$^{1}$Department of Computer Science and Technology, Tsinghua University, Beijing, China \\
$^{2}$Beijing National Research Center for Information Science and Technology\\
$^{3}$Institute for Artificial Intelligence, Tsinghua University, Beijing, China\\
$^{4}$Department of Electronic Engineering, Tsinghua University, Beijing, China \\
{\tt \{wh-li20, qfc17, yi-xy16, zhangjr18\}@mails.tsinghua.edu.cn}\\
{\tt sms@tsinghua.edu.cn}
}

\begin{document}
\maketitle
\begin{abstract}
Poetry is one of the most important art forms of human languages. Recently many studies have focused on incorporating some linguistic features of poetry, such as style and sentiment, into its understanding or generation system. 
However, there is no focus on understanding or evaluating the semantics of poetry. 
Therefore, we propose a novel task to assess a model's semantic understanding of poetry by poem matching. 
Specifically, this task requires the model to select one line of Chinese classical poetry among four candidates according to the modern Chinese translation of a line of poetry.
To construct this dataset, we first obtain a set of parallel data of Chinese classical poetry and modern Chinese translation. Then we retrieve similar lines of poetry with the lines in a poetry corpus as negative choices. 
We name the dataset Chinese Classical Poetry Matching Dataset (CCPM) and release it at \url{https://github.com/THUNLP-AIPoet/CCPM}. We hope this dataset can further enhance the study on incorporating deep semantics into the understanding and generation system of Chinese classical poetry. 
We also preliminarily run two variants of BERT \citep{devlin-etal-2019-bert} on this dataset as the baselines for this dataset.
\end{abstract}

\section{Introduction}

Language is one of the most crucial forms of human intelligence. Among all the genres of human language, poetry is a distinctive artistic genre with exquisite expression, rich content, and diverse styles. In the long history of humankind, poetry shows profound impacts across different countries, nationalities, and cultures. 

Poetry has various distinguishing characteristics from other genres, including powerful emotion, explicit language style, and rich content expressed in an abstractive manner. These characteristics differentiate the automatic processing of poetry from the processing of other genres by a large margin. As a result, there have been many works focusing on some of their features of poetry such as style \citep{yang2018stylistic, liu2020construction} and sentiment \citep{chen2019sentiment}. However, to our best knowledge, there is no work concentrating on the internal semantics of poems. 
There may be a possible reason. In poem writing, the poet often needs to compress plentiful meanings into the limited length of contents constrained by the genre. Therefore, the semantics presented in the poetry is much fuzzier and more entangled among different segments than other genres. That leads to the difficulty of automatically analyzing and evaluating poem semantics, which encourages more work in this area. 

\begin{figure}[t]
    \centering
    \includegraphics[width=\linewidth]{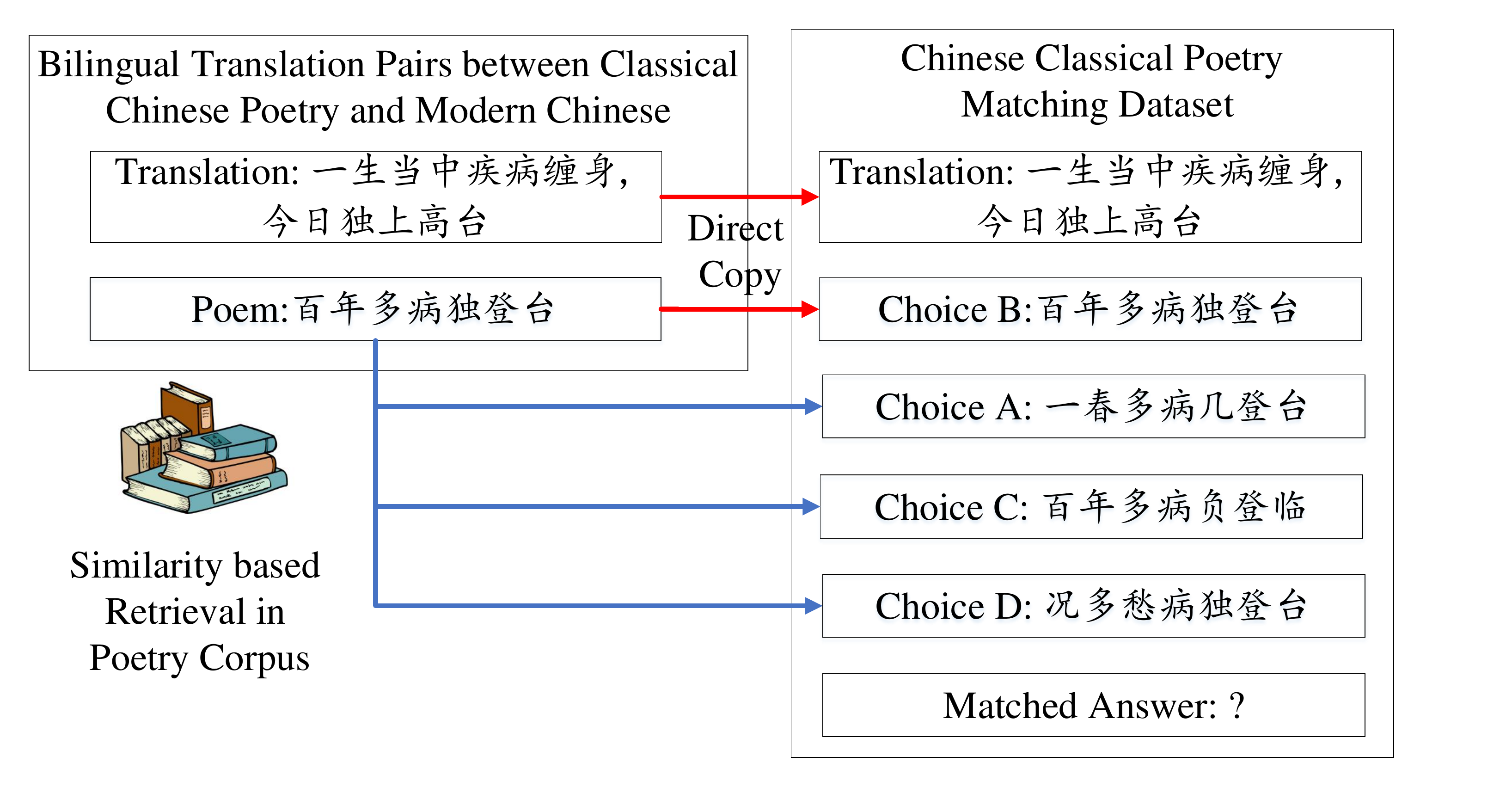}
    \caption{The major process of the dataset construction. We first collect bilingual translation pairs and retrieve the most similar candidates from the poetry corpus as confusing choices}
    \label{fig:my_label}
\end{figure}

Therefore, in this work, we propose a benchmark on the semantics of Chinese classical poems. More specifically, we design a novel task to quantify the semantic modeling ability around different models by poetry matching. This task is to test whether the model can discern the correct poem line with other similar lines given the corresponding translation of the correct line in the modern Chinese.

Meanwhile, we established a dataset for this task. 
We first collected 31K pairs of bilingual parallel data between Chinese classical poems and modern Chinese. 
Then we cleaned the data and retrieved the most similar poem lines in our poetry corpus for each poem line.
The major process of constructing this dataset is showed in Section ~\ref{sec:cons}.

We hope this dataset can further enhance the research on semantics in poetry. It can benefit the semantic understanding of the poetry analysis models. It also bridges the semantics of modern Chinese with uncommonly used poetic language, providing a chance for poetry generation models to better understand the users' intend and improve the semantic relevance between the user input and the generated poem.

Therefore, the contributions of this work lie in:
\begin{itemize}
    \item Propose a new task to match the translation of Chinese classical poems in modern Chinese to its original poem lines;
    \item Release a dataset on this task to further evaluate and improve the semantic understanding of both automatic analysis and automatic generation models of Chinese classical poems.
\end{itemize}

\section{Related Work}
There have been several open datasets about Chinese classical poetry. The first fine-grained dataset, to our best knowledge, is released by \citet{chen2019sentiment}. They annotated a manually-labeled fine-grained Sentiment Poetry Corpus including 5,000 Chinese quatrains. They also released a Chinese poetry dataset CCPC, which contains 151,835 unlabelled Chinese quatrains. Moreover, \citet{liu2020construction} released a dataset with 3940 quatrains with themes and 1917 with sentiments automatically annotated by template-based methods. They also constructed a knowledge graph of Chinese classical poetry using the Apriori algorithm and human-defined scheme system. To our best knowledge, the most similar work to ours is \citet{liu2019ancient}. 
They collected a dataset of the parallel bilingual pairs between ancient Chinese and modern Chinese from the web. They also used string matching algorithms to align the lines in parallel pairs. However, they focus on using the data to train a neural machine translation model to tackle the poetry translation poem. 
Instead, we aim to construct a matching dataset based on those bilingual pairs to assess the semantic understanding ability of the models.

\section{Dataset Construction}
\label{sec:cons}
\subsection{Bilingual Pair Extraction}
The extraction of bilingual data mainly consists of three subprocesses: raw data acquiring, line segmenting, and format filtering.

\paragraph{Raw data acquirement} We collected 6K paragraphs in ancient Chinese and their corresponding translation in modern Chinese language from the web to construct this dataset. They include Chinese classical poetry and other literature in the ancient Chinese language. 

\paragraph{Line segmentation}  As stated in Introduction, we want to build our dataset on the level of poem lines. Therefore, we need to conduct line segmentation. 
We split some instances into lines by the line separation in the raw web contents and punctuation. For some other instances, we can only split them into the units of two lines. Considering that this unit can also have consecutive semantics, we also keep this part of examples in our dataset in the form of two-line instances.

\paragraph{Format filtering} 
We only want to keep Chinese classical poems in our dataset, so we need to filter the others out. Since the 5-character line (5 yan) poems and the 7-character (7 yan) line poems are the most common ones, we only keep lines in these two formats in our dataset. We achieve this by specifying a length constraint on the two consecutive lines.

After these steps, we extracted 27,218 bilingual pairs of Chinese classical poetry and their translations into modern Chinese for our dataset. For more detailed statistics, please refer to Section \ref{sec:stat}.

\subsection{Candidates Retrieval}

After obtaining the bilingual pairs, the next step is to extract the negative choices according to the correct answer. To select more confusing opposing candidates, we need to seek more similar poem lines with the ground truth answer. Therefore, we used our pre-trained model on poetry, BERT-CCPoem \cite{CCPOEM}, to calculate the similarity between sentences. This model is trained on an almost complete collection of Chinese classical poems, CCPC-Full v1.0, consisting of 926,024 classical poems with 8,933,162 sentences.

We use the hidden state of the \texttt{[CLS]} token as the semantic representation of the whole sentence and use the cosine similarity between two lines' representations as the similarity metric between them. We also adopt the Locality Sensitive Hashing (LSH) algorithm \cite{arya1998optimal} to speed up retrieving the most similar sentences. The top candidates retrieved are re-ranked by the weighted average of the BERT similarity and the Longest Common Sequence (LCS) between each candidate and the ground truth poem line like in Equation ~\eqref{eq1}. More details of this re-ranking algorithm can be seen in \citet{guo2020ranking}.
\begin{equation}
\begin{aligned}
\operatorname{Sim}(A, B) =& ~ 0.4 * \operatorname{LCS}(A, B)\\
&+ 0.6 * \operatorname{EMB}(A, B).
\end{aligned}
\label{eq1}
\end{equation}
After choosing the most similar candidates, we select the most similar candidate and randomly sample two choices separately from the top 2-5 and the top 6-10 similar candidates. We use these three as the negative choices and shuffle them with the ground truth answer.

\subsection{Dataset Statistics}
\label{sec:stat}
We obtain 27,218 translation-candidate pairs and split them into the training, validation, and test sets. The detailed statistics are in Table ~\ref{tab:stat}.


\begin{table}[t]
\centering
\resizebox{\linewidth}{!}{%
\begin{tabular}{@{}c|rrr|r@{}}
\toprule
 & \multicolumn{1}{c}{Training} & \multicolumn{1}{c}{Validation} & \multicolumn{1}{c|}{Test} & \multicolumn{1}{c}{Total} \\ \midrule
5-yan 1-line  & 10,166 & 1,270 & 1,270 & 12,706 \\
7-yan 1-line  & 9,392  & 1,173 & 1,173 & 11,738 \\
5-yan 2-lines & 1,025  & 128   & 128   & 1,281  \\
7-yan 2-lines & 1,195   & 149   & 149   & 1,493  \\ 
\hline
Total         & 21,778 & 2,720 & 2,720 & 27,218 \\ 
\bottomrule
\end{tabular}%
}
\caption{Detailed dataset statistics: \textit{5-yan} and \textit{7-yan} refer to 5 or 7 Chinese characters in one poem line, and \textit{1-line} and \textit{2-lines} refer to 1 or 2 lines in one sample.}
\label{tab:stat}
\end{table}

\section{Experiments}
In this section, we evaluate some popular NLP models on the CCPM dataset.

\paragraph{Evaluation Metric}
We use accuracy, namely the percentage of the test samples for which the evaluated model predicts the correct answer, as the evaluation metric.
The higher the accuracy is, the model performs better.

\paragraph{Evaluated Models}
We choose the popular pre-trained language model BERT \citep{devlin-etal-2019-bert} as the sentence encoder and design the following two models:

(1) BERT-Cls. Similar to previous work \citep{devlin-etal-2019-bert}, it concatenates the given translation with each candidate poem line (with an additional separator token) and feeds the concatenation into BERT. Then it feeds the hidden state of the \texttt{[CLS]} token into 
a linear layer and makes a binary classification: match or not.

(2) BERT-Match, which regards the task as a sentence match problem. Specifically, it uses two BERTs to encode the translation and candidate poem lines, respectively, obtaining the embeddings of translation and poem lines. Then it calculates the cosine similarity between the embeddings of translation and each poem line. The candidate poem that is most similar to the translation is selected as the final answer.

\paragraph{Implementation Details}
For both models, we choose \texttt{bert-base-chinese} from the Transformer library \citep{wolf2020transformers} as the sentence encoder.
During fine-tuning, we use the Adam optimizer \citep{kingma2014adam}, with an initial learning rate $2e-5$ that decreases linearly and train the models for $4$ epochs.

\begin{table}[t]
\centering
\resizebox{.6\linewidth}{!}{%
\begin{tabular}{cc}
\toprule
Model & Accuracy \\
\midrule
BERT-Cls & 84.96 \\
BERT-Match & 82.60 \\
\bottomrule
\end{tabular}%
}
\caption{The accuracy results of the evaluated model on the test set of CCPM.}
\label{tab:results}
\end{table}

\paragraph{Results}
The accuracy results of the two models are shown in Table \ref{tab:results}.
We can see that the two models perform similarly and both achieve an acceptable accuracy.
BERT-Cls outperforms BERT-Match, 
presumably because BERT-Cls achieves more interaction between the translation and a poem line with the self-attention mechanism of Transformer \citep{vaswani2017attention} of BERT. 

\section{Conclusion}
This paper proposed a novel task of poem matching, which is aimed at assessing the semantic understanding ability to Chinese classical poetry. Further, we constructed a dataset for this task by collecting the parallel data of the classical poems and their translations and retrieving similar poem lines in the poetry corpus as confusion choices. We also ran two variants of the BERT models and compared the results.

In the future, we will further refine this dataset in three ways. Firstly, we will collect more parallel data to increase the volume of this dataset. Secondly, we will explore more confusing ways to construct negative choices. Thirdly, we will test more commonly used NLP models on our benchmark.

\bibliography{custom}
\bibliographystyle{acl_natbib}

\end{document}